%% file: main.tex
\documentclass[10pt,twocolumn,letterpaper]{article}

\usepackage[pagenumbers]{cvpr} 

\usepackage[table]{xcolor}

\newcommand{\invis}[1]{\textcolor{red}{#1}}
\newcommand{\vis}[1]{\textcolor{green!60!black}{#1}}

\definecolor{cvprblue}{rgb}{0.21,0.49,0.74}
\usepackage[pagebackref,breaklinks,colorlinks,allcolors=cvprblue]{hyperref}

\def\paperID{*****} 
\def\confName{CVPR}
\def\confYear{2026}

\title{Hand Visibility Detector: Per-Keypoint Visibility Estimation for Hands}

\author{Ryosei Hara$^{1,2}$, 
Masashi Hatano$^{4}$, 
Rintaro Yanagi$^{2}$,
Atsushi Hashimoto$^{3}$,
Takuma Yagi$^{2}$,
Mariko Isogawa$^{1,2}$
\\[0.5em]
$^{1}$Keio University,
$^{2}$National Institute of Advanced Industrial Science and Technology (AIST), \\
$^{3}$OMRON SINIC X Corporation,
$^{4}$The University of Tokyo 
}

\begin{document}
\maketitle

\begin{abstract}
Hand Pose Estimation (HPE) is a fundamental technology for various applications such as AR/VR and robotics.
In these applications, the visibility of each hand joint in the image is crucial for assessing the reliability of estimation results under occlusion.
However, most existing HPE methods output joint positions without explicitly indicating their visibility.
Although some methods account for occlusion or visibility, visibility estimation has mainly been used as an auxiliary signal for improving pose estimation.
To our knowledge, per-joint hand visibility estimation has not been systematically studied as a standalone task.
In this work, we propose Hand Visibility Detector, a model for estimating the visibility of individual hand joints, and present the first systematic investigation of visibility estimation as an independent task.
We show that leveraging the prior knowledge of HPE models pretrained on large-scale data as a backbone yields high performance in this task.
We further demonstrate the utility of Hand Visibility Detector on a downstream task of 3D hand pose annotation via multi-view triangulation of 2D keypoints, showing that visibility-weighted triangulation reduces reprojection error.
Our method is released as a ready-to-use package, and the code and demo are available at
\url{https://github.com/ryhara/hand_visibility_detector}.
\end{abstract}

\vspace{-2mm}
\section{Introduction}
\label{sec:intro}

\begin{figure}[t]
  \centering
  \includegraphics[width=0.75\linewidth]{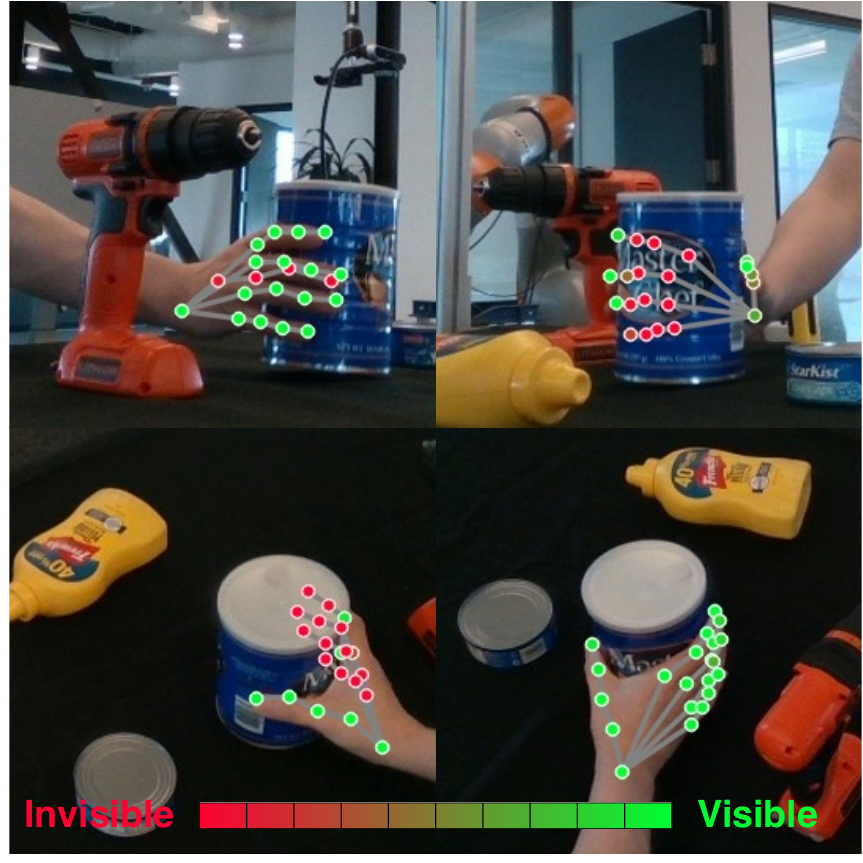}
  \caption{\textbf{Overview of Hand Visibility Detector.} Our model estimates per-joint visibility as a confidence score in $[0, 1]$. \invis{Red} indicates invisible joints, and \vis{green} indicates visible joints.}
  \label{fig:teaser}
  \vspace{-2mm}
\end{figure}

Hand poses play an important role in various fields, including AR/VR, human-computer interaction, and robotics~\cite{Ohkawa2023survey}.
Hand Pose Estimation (HPE) from images has advanced significantly with deep learning, and large-scale pretrained models that achieve highly accurate estimation even from monocular images have recently emerged, driven by Vision Transformers~\cite{dosovitskiy2021vit} and the scaling of training data~\cite{pavlakos2024hamer, potamias2025wilor}.

However, most existing HPE methods always output joint positions without explicitly indicating their visibility.
Visibility refers to whether each joint is directly observable in the image, distinguishing observable joints from those whose locations must be inferred under occlusion.
Several HPE methods explicitly model occlusion or visibility to improve pose estimation~\cite{Ye_2018_ECCV, kim2021endtoend, park2022handoccnet, xu2023h2onet}, as in other tasks such as point tracking~\cite{harley2022particle, karaev2024cotracker} and hand forecasting~\cite{hatano2025invisible}.
However, visibility is typically treated as an auxiliary signal, and the quality of the visibility estimation itself is not the primary object of study.
More recently, Contact4D~\cite{song2026contact4d} has utilized per-joint visibility to refine triangulated 3D hand keypoints for occlusion-robust automatic annotation from multi-view images, indicating that visibility can be an informative cue in its own right, beyond its conventional role as an auxiliary component.
However, because visibility is primarily evaluated indirectly through downstream pose accuracy, it remains unclear how accurately these models predict joint visibility itself and how well such predictions generalize across various occlusion conditions.

In this work, we propose Hand Visibility Detector, a model dedicated to hand joint visibility estimation.
Our method adopts a simple architecture that attaches a lightweight visibility head to the frozen backbone of an HPE model~\cite{pavlakos2024hamer, potamias2025wilor} pretrained on large-scale hand data.
For training and evaluation, we use the HInt dataset~\cite{pavlakos2024hamer}, which provides manual per-joint visibility annotations on diverse in-the-wild data, including web images and egocentric video frames.
Unlike prior work, which was trained and evaluated only on a single domain, the diversity of this dataset allows us to demonstrate the effectiveness of the prior knowledge of pretrained HPE models for visibility estimation under diverse conditions.
Furthermore, to demonstrate its utility for automatic 3D annotation, we confirm that visibility-weighted triangulation of multi-view 2D keypoints reduces reprojection errors on three multi-view datasets for 3D hand pose estimation, DexYCB~\cite{chao2021dexycb}, HO3D~\cite{hampali2020ho3d}, and H2O~\cite{kwon2021h2o}.

Our contributions are summarized as follows:
\begin{itemize}
    \item We formulate per-joint hand visibility estimation as a standalone task and introduce Hand Visibility Detector together with a systematic evaluation of this task.
    \item Our method outperforms baseline methods by 3.4 points in mAP on the HInt dataset, demonstrating the effectiveness of the prior knowledge of large-scale pretrained HPE models for visibility estimation.
    \item We evaluate visibility-weighted triangulation as a downstream task, reducing the mean reprojection error by up to 10.1\% on DexYCB, HO3D, and H2O.
\end{itemize}

\section{Related Work}
\label{sec:related}

\subsection{Hand Pose Estimation}
\label{ssec:hand_pose}
Recent HPE methods typically employ deep learning to estimate the parameters of the parametric MANO hand model~\cite{romero2017MANO}.
Boukhayma et al.~\cite{Boukhayma_2019_CVPR} pioneered direct regression of MANO parameters from a monocular image, followed by regression-based methods such as METRO~\cite{lin2021metro} and Mesh Graphormer~\cite{lin2021graphormer}.
Occlusion-aware HPE has also been actively studied to handle occlusions caused by objects~\cite{park2022handoccnet, lin2023hflnet, xu2023h2onet} and two-hand interactions~\cite{hampali2022keypoint}.
More recently, Vision Transformers and the scaling of training data have led to large-scale pretrained models such as HaMeR~\cite{pavlakos2024hamer} and WiLoR~\cite{potamias2025wilor}, which achieve remarkable accuracy even on in-the-wild images.
In this work, we transfer the rich prior knowledge of these pretrained HPE models to a new task: hand joint visibility estimation.

\subsection{Hand Joint Visibility Detection}
\label{ssec:visibility}
While no prior work has addressed hand joint visibility estimation as an independent task, several methods estimate visibility in the course of HPE.
Ye et al.~\cite{Ye_2018_ECCV} proposed a hierarchical mixture density network for hand pose estimation from depth images, which models per-joint visibility as a Bernoulli distribution and assigns a unimodal Gaussian distribution to visible joints and a multimodal Gaussian mixture to occluded ones.
Kim et al.~\cite{kim2021endtoend} introduced a per-joint visibility estimator into end-to-end estimation of two interacting hands and utilized its output for visibility-guided 2D joint heatmap enhancement.
As non-per-joint approaches, HDR~\cite{meng2022hdr} recovers the occluded appearance of the target hand and removes the occluding hand via amodal segmentation and appearance recovery, and H2ONet~\cite{xu2023h2onet} estimates finger-level occlusion probabilities to weight multi-frame feature aggregation.
Since all of these methods are trained on datasets captured in controlled environments, their generalization ability as visibility estimators has not been validated.
In this work, we use the HInt dataset~\cite{pavlakos2024hamer}, which provides manual per-joint visibility annotations on diverse in-the-wild data.
To the best of our knowledge, this is the first systematic evaluation of the performance and generalization of visibility estimation itself.

\subsection{Automatic 3D Hand Pose Annotation}
\label{ssec:annotation}
Acquiring accurate 3D annotations is essential for 3D human body and hand pose estimation research~\cite{Ohkawa2023survey}.
Optical markers~\cite{motive} and glove-based motion capture~\cite{kim2025parahome} are accurate but require dedicated equipment, and the attached devices compromise the natural appearance of hands.
Triangulation from multi-view images~\cite{chao2021dexycb, hampali2020ho3d, kwon2021h2o, ohkawa2023assemblyhands} is therefore widely used, but its accuracy depends on the 2D HPE accuracy in each view.
Standard triangulation treats all views equally, and even when outliers are removed with RANSAC~\cite{moon2020interhand, wang2025hocap}, the remaining low-confidence points are used as they are.
Contact4D~\cite{song2026contact4d} achieves occlusion-robust annotation by weighting triangulation with the detection confidence of a hand detector~\cite{potamias2025wilor} and refining the triangulated 3D joint positions using a visibility estimator, based on RTMPose~\cite{jiang2023rtmposerealtimemultipersonpose} (with a CSPNeXt~\cite{CHEN2024cspnext} backbone) and trained on COCO-WholeBody~\cite{jin2020whole}.
However, the performance of this visibility estimator itself has not been validated. Moreover, we observed that the hand visibility labels in COCO-WholeBody are often inaccurate or missing, which motivates us to use a higher-quality label source instead.
In this work, we train a visibility estimator on the HInt dataset and demonstrate that visibility-weighted triangulation using our Hand Visibility Detector reduces reprojection errors on three multi-view hand datasets, DexYCB~\cite{chao2021dexycb}, HO3D~\cite{hampali2020ho3d}, and H2O~\cite{kwon2021h2o}, compared to unweighted and hand detection confidence-weighted baselines.

\vspace{-2mm}
\section{Method}
\label{sec:method}

\begin{figure*}[t]
  \centering
  \includegraphics[width=\linewidth]{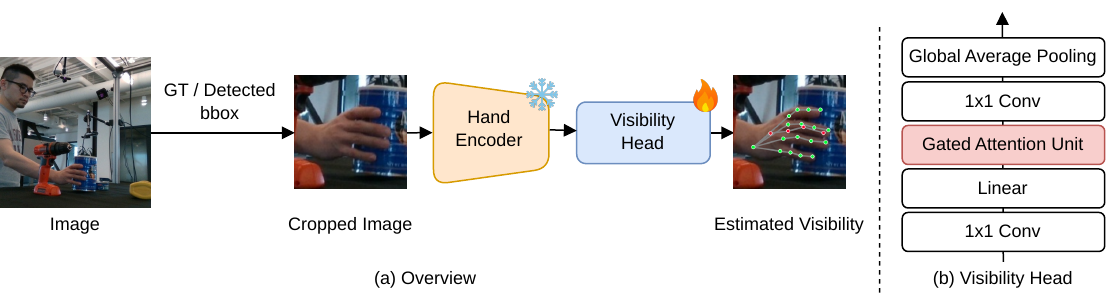}
  \vspace{-8mm}
  \caption{\textbf{Overview of the proposed method.} (a) Overall pipeline. Our model outputs only per-joint visibility, shown as \vis{green} (visible) and \invis{red} (invisible). Joint positions are provided by ground truth or an off-the-shelf HPE model for visualization only. (b) Architecture of the Visibility Head. A frozen backbone of a pretrained HPE model extracts features, and only the lightweight head is trained.}
  \label{fig:method}
  \vspace{-2mm}
\end{figure*}

\cref{fig:method} shows an overview of the proposed method.
Given a hand-cropped image $I \in \mathbb{R}^{H \times W \times 3}$, our method estimates per-joint visibility $V = (\hat{v}_1, \ldots, \hat{v}_J) \in [0,1]^{J}$.
We use ground-truth boxes provided by the datasets for training and evaluation, and boxes obtained by the WiLoR~\cite{potamias2025wilor} hand detector in the downstream task.
The joint layout follows the 21 keypoints of the MANO hand model~\cite{romero2017MANO} ($J=21$), and each $\hat{v}_j$ represents the probability that joint $j$ is visible.
Our method consists of (i) a frozen ViT backbone of a pretrained HPE model and (ii) a lightweight visibility head.

\subsection{Hand Encoder}
\label{ssec:backbone}
Since occluded joints lack direct visual evidence at their location, determining their visibility requires reasoning about the entire hand structure, such as its spatial relationships with visible fingers and potential occluders.
To perform such reasoning across diverse scenes, it is desirable to leverage rich prior knowledge of hand structure, rather than training a feature extractor from scratch on the limited visibility labels of a specific dataset as in prior work~\cite{Ye_2018_ECCV, kim2021endtoend}.
Large-scale pretrained HPE models such as HaMeR~\cite{pavlakos2024hamer} and WiLoR~\cite{potamias2025wilor} have learned to recover 3D poses from millions of diverse hand images under various conditions, including occlusion, and are thus expected to have acquired exactly the prior knowledge of hand structure necessary for reasoning about occluded joints.

We therefore adopt the pretrained ViT backbones of HaMeR and WiLoR as our Hand Encoder.
Given the input image $I$, the encoder produces a feature map $F \in \mathbb{R}^{h \times w \times C}$ by rearranging the output tokens into their spatial layout.
The backbone parameters are frozen at their pretrained values.
This preserves the prior knowledge of the large-scale pretrained model while avoiding overfitting to the relatively small visibility-labeled data.

\subsection{Visibility Head}
\label{ssec:head}
To estimate per-joint visibility from the feature map $F$, we employ a lightweight visibility head adopting the design of the head architecture of RTMPose~\cite{jiang2023rtmposerealtimemultipersonpose}, which is also adopted by the visibility estimator of Contact4D~\cite{song2026contact4d}.
Specifically, a $1 \times 1$ convolution first compresses the feature map from $C$ to $d$ dimensions, which is then flattened into a sequence of $h \times w$ tokens.
After a fully connected layer, a Gated Attention Unit (GAU)~\cite{pmlr-v162-hua22a} models global dependencies among spatial positions.
GAU is an attention mechanism that integrates self-attention with gated linear units, and its effectiveness for keypoint estimation has been shown in RTMPose.
Finally, a $1 \times 1$ convolution projects the features into $J$ channels, spatial average pooling produces per-joint logits, and a sigmoid function converts them into visibility probabilities $\hat{v}_j$.

We train the head with the binary cross-entropy loss against ground-truth visibility labels $v_j \in \{0,1\}$:
\vspace{-1mm}
\begin{equation}
  \mathcal{L} = -\frac{1}{J}\sum_{j=1}^{J} \left[ v_j \log \hat{v}_j + (1-v_j) \log (1-\hat{v}_j) \right],
  \label{eq:loss}
\end{equation}
where joints outside the frame are treated as invisible ($v_j = 0$).
That is, visibility in this work means that a joint is neither occluded nor outside the frame.
We freeze the backbone and train only the visibility head, allowing visibility estimation to be learned at an extremely low cost while preserving the prior knowledge of the pretrained HPE model.

\section{Experiments}
\label{sec:experiment}

\noindent\textbf{Dataset.}
We train and evaluate on the HInt dataset~\cite{pavlakos2024hamer}, which provides manually annotated keypoints with per-joint visibility labels.
HInt covers diverse in-the-wild scenes, including web and egocentric videos, and we use 25,273 frames for training and 5,374 for evaluation.

\noindent\textbf{Baseline Methods.}
We compare with the following two visibility estimators from existing methods capable of per-joint visibility estimation.
The visibility estimator of Kim et al.~\cite{kim2021endtoend} consists of a ResNet-50~\cite{he2016resnet} pretrained on ImageNet-1k~\cite{Russakovsky2015imagenet} and a two-layer linear head.
The visibility estimator of Contact4D~\cite{song2026contact4d} follows RTMPose~\cite{jiang2023rtmposerealtimemultipersonpose} and uses a CSPNeXt~\cite{CHEN2024cspnext} backbone pretrained on ImageNet-1k.

\noindent\textbf{Evaluation Metrics.}
We evaluate visibility estimation as a per-joint binary classification problem.
We use mAP (average precision averaged over joints) and F1 score, where F1 is computed by binarizing predictions at a threshold of 0.5.
All experiments are run with three seeds, and we report the mean and standard deviation.

\noindent\textbf{Implementation Details.}
The input image is obtained by expanding the ground-truth hand bounding box by a factor of 1.25, resizing it to $256 \times 256$, and cropping the central $H \times W = 256 \times 192$  region following WiLoR.
The encoder divides the input into $16 \times 16$ patches, yielding a feature map of $h=16$, $w=12$, and $C=1280$.
The hidden dimension of the visibility head is $d=256$ with a dropout rate of 0.1.
Only the visibility head is trained, and its 0.83M parameters account for merely 0.131\% of the 631M-parameter model.
We use AdamW~\cite{Loshchilov2017DecoupledWD} with a learning rate of $1.0 \times 10^{-3}$.
We train for 100 epochs with a batch size of 256, which takes about 2.5 hours on a single NVIDIA H200 GPU using approximately 10GB of GPU memory.
All experiments are conducted under the same settings.
For the backbone ablation, each backbone is initialized with its official pretrained checkpoint and kept frozen.
Please refer to our released code for further details.

\section{Results}
\label{sec:result}

\subsection{Visibility Estimation}
\label{ssec:result_visibility}

\begin{table}[t]
  \centering
  \caption{\textbf{Comparison with baseline methods on HInt.}}
  \vspace{-2mm}
  \label{tab:comparison}
  \small
  \setlength{\tabcolsep}{4pt}
  \begin{tabular}{lcc}
    \toprule
    Method  & mAP $(\uparrow)$ & F1 $(\uparrow)$ \\
    \midrule
    Kim et al.\ \cite{kim2021endtoend}  & 0.895\,{\scriptsize$\pm 0.001$} & 0.858\,{\scriptsize$\pm 0.001$}  \\
    Contact4D \cite{song2026contact4d}  & 0.897\,{\scriptsize$\pm 0.001$} & 0.860\,{\scriptsize$\pm 0.002$} \\
    \rowcolor{gray!20}
    Ours                                & \textbf{0.931}\,{\scriptsize$\pm0.000$} & \textbf{0.896}\,{\scriptsize$\pm0.001$}  \\
    \bottomrule
  \end{tabular}
  \vspace{-2mm}
\end{table}

\begin{table}[t]
  \centering
  \caption{\textbf{Ablation on backbones.} The visibility head and hyperparameters are shared across all models.}
  \vspace{-2mm}
  \label{tab:backbone}
  \resizebox{\linewidth}{!}{%
  \begin{tabular}{lrcc}
    \toprule
    Backbone & \#Params & mAP $(\uparrow)$ & F1 $(\uparrow)$ \\
    \midrule
    CSPNeXt-X \cite{CHEN2024cspnext}    &  48M & $0.800$\,{\scriptsize$\pm0.002$} & $0.799$\,{\scriptsize$\pm0.001$} \\
    ResNet-152 \cite{he2016resnet}     &  58M & $0.796$\,{\scriptsize$\pm0.001$} & $0.797$\,{\scriptsize$\pm0.001$} \\
    ViT-H \cite{dosovitskiy2021vit} & 633M & $0.838$\,{\scriptsize$\pm0.005$} & $0.819$\,{\scriptsize$\pm0.004$} \\
    DINOv3 \cite{simeoni2026dinov3}    & 840M & $0.897$\,{\scriptsize$\pm0.000$} & $0.866$\,{\scriptsize$\pm0.000$} \\
    \midrule
    \rowcolor{gray!20}
    HaMeR \cite{pavlakos2024hamer}     & 631M & $0.932$\,{\scriptsize$\pm0.000$} & ${0.896}$\,{\scriptsize$\pm0.000$} \\
    \rowcolor{gray!20}
    WiLoR \cite{potamias2025wilor}     & 631M & $0.931$\,{\scriptsize$\pm0.000$} & $0.896$\,{\scriptsize$\pm0.001$} \\
    \midrule
    WiLoR (fine-tuned) \cite{potamias2025wilor}   & 631M & 0.622 \,{\scriptsize$\pm0.014$} & 0.704 \,{\scriptsize$\pm0.018$} \\
    \bottomrule
  \end{tabular}%
  }
  \vspace{-2mm}
\end{table}

\noindent\textbf{Comparison with Baseline Methods.}
\cref{tab:comparison} shows the comparison with existing methods capable of visibility estimation.
Our method achieves an mAP of 0.931 and an F1 of 0.896, substantially outperforming Kim et al.~\cite{kim2021endtoend} and Contact4D~\cite{song2026contact4d} on both metrics.
Both baselines rely on CNN features pretrained on ImageNet-1k, and this gap demonstrates the effectiveness of the hand prior knowledge of pretrained HPE models.

\noindent\textbf{Choice of Backbone.}
\cref{tab:backbone} shows the visibility estimation performance of each backbone.
The hand-specific models HaMeR and WiLoR perform best, and even DINOv3, the strongest among recent general-purpose image models, falls short of them.
Moreover, fine-tuning the WiLoR backbone for visibility estimation degrades mAP from 0.931 to 0.622.
This indicates that task-specific fine-tuning corrupts the feature representations acquired during pre-training, supporting our design of freezing the backbone and training only the head.

\begin{table}[t]
  \centering
  \caption{\textbf{Ablation on the visibility head.} The backbone is fixed to our frozen WiLoR encoder.}
  \vspace{-2mm}
  \label{tab:ablation}
  \small
  \resizebox{\linewidth}{!}{
  \begin{tabular}{lcc}
    \toprule
    Head  & mAP $(\uparrow)$ & F1 $(\uparrow)$ \\
    \midrule
    w/ Kim et al. head~\cite{kim2021endtoend} & 0.905\,{\scriptsize$\pm0.000$} & 0.874\,{\scriptsize$\pm0.001$} \\
    w/o GAU              & 0.887\,{\scriptsize$\pm0.000$} & 0.860\,{\scriptsize$\pm0.000$}  \\
    \rowcolor{gray!20}
    Ours (full)          & \textbf{0.931}\,{\scriptsize$\pm0.000$} & \textbf{0.896}\,{\scriptsize$\pm0.001$} \\
    \bottomrule
  \end{tabular}}
  \vspace{-2mm}
\end{table}

\noindent\textbf{Design of Visibility Head.}
\cref{tab:ablation} shows the ablation on the visibility head with the backbone fixed to WiLoR.
Removing the GAU degrades mAP from 0.931 to 0.887, and replacing our head with the linear head of Kim et al.~\cite{kim2021endtoend} also falls short of ours.
This demonstrates that the GAU, which models global dependencies among spatial positions, is effective for visibility estimation.

\begin{figure}[t]
  \centering
  \includegraphics[width=0.75\linewidth]{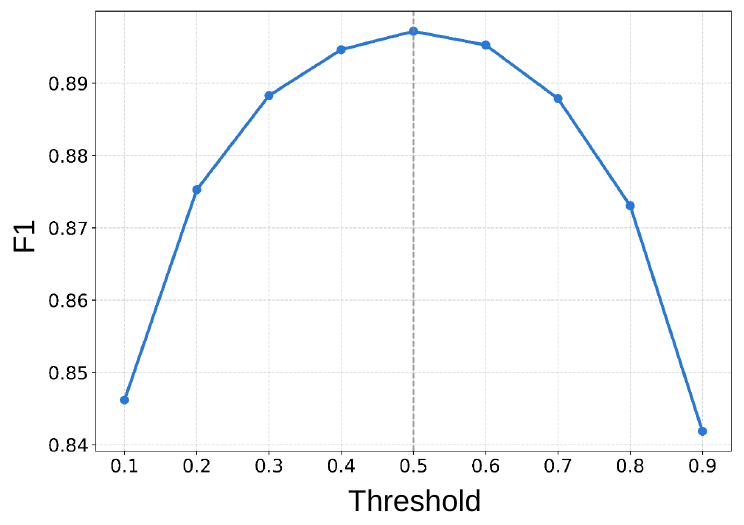}
  \vspace{-2mm}
  \caption{\textbf{F1 score versus binarization threshold.} Performance peaks around 0.5 and remains stable over a wide range.}
  \label{fig:threshold}
  \vspace{-2mm}
\end{figure}

\noindent\textbf{Visibility Threshold Sensitivity.}
\cref{fig:threshold} shows the F1 score as the binarization threshold varies from 0.1 to 0.9.
F1 peaks around a threshold of 0.5 and stays above 0.88 over the wide range of 0.3--0.7, indicating that performance is relatively insensitive to the threshold.

\begin{figure*}[t]
  \centering
  \includegraphics[width=0.9\linewidth]{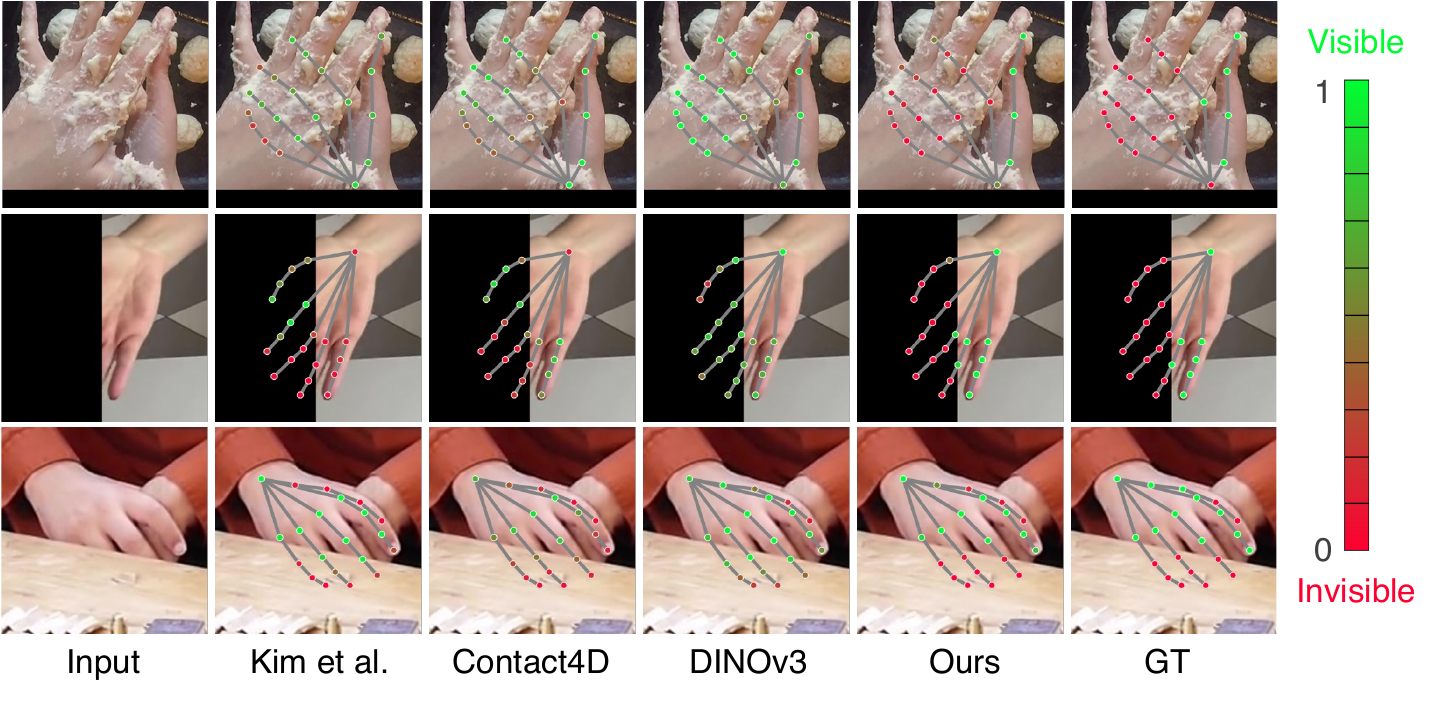}
  \vspace{-4mm}
  \caption{\textbf{Qualitative results of visibility estimation.} Joint positions are ground truth. \vis{Green} indicates joints predicted as visible, and \invis{red} indicates joints predicted as invisible. From top to bottom: self-occlusion, image truncation, and occlusion by objects.}
  \label{fig:qualitative}
  \vspace{-2mm}
\end{figure*}

\noindent\textbf{Qualitative Evaluation.}
\cref{fig:qualitative} shows qualitative comparisons with the baselines and the backbone ablation.
Our method correctly estimates visibility with the highest confidence in all cases of self-occlusion, image truncation, and occlusion by objects.

\subsection{Downstream Task: Triangulation}
\label{ssec:result_triangulation}
\noindent\textbf{Settings.}
As a downstream task, we conduct a triangulation experiment assuming automatic 3D hand pose annotation from multi-view images.
We triangulate 2D keypoints estimated by WiLoR~\cite{potamias2025wilor} in each view using DLT~\cite{ABDELAZIZ2015dlt}, and compare three weighting schemes:
(i) unweighted triangulation (Baseline),
(ii) weighting by the WiLoR detector confidence, shared by all joints in a view~\cite{song2026contact4d} (Detection conf.\ weighted), and
(iii) weighting by the per-joint visibility estimated by our method (Visibility weighted).
We evaluate reprojection errors (px) on the test sets of three datasets: DexYCB~\cite{chao2021dexycb} (12,902 frames, 8 views, $640 \times 480$), HO3D~\cite{hampali2020ho3d} (4,854 frames, 2 or 5 views, $640 \times 480$), and H2O~\cite{kwon2021h2o} (23,391 frames, 4 views, $1280 \times 720$).

\begin{figure}[t]
  \centering
  \includegraphics[width=\linewidth]{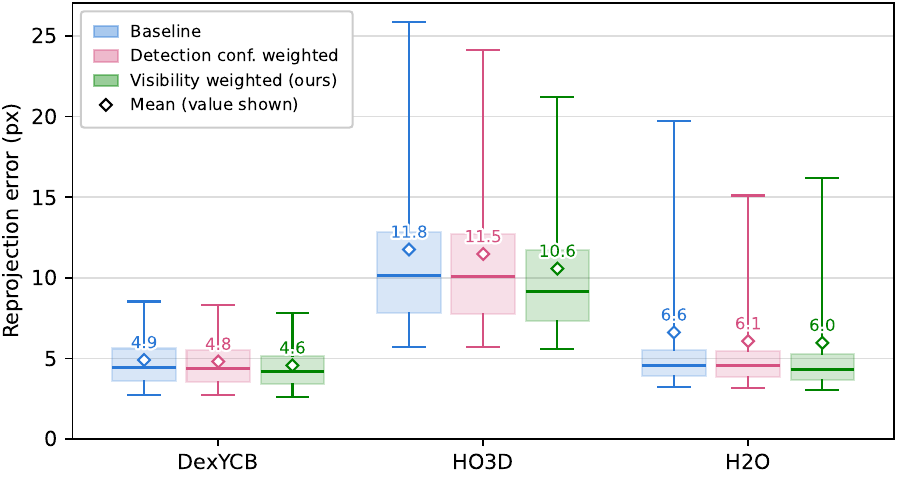}
  \vspace{-6mm}
  \caption{\textbf{Reprojection errors of triangulation on each dataset.}}
  \label{fig:triangulation}
  \vspace{-4mm}
\end{figure}

\begin{figure}[t]
  \centering
  \includegraphics[width=0.8\linewidth]{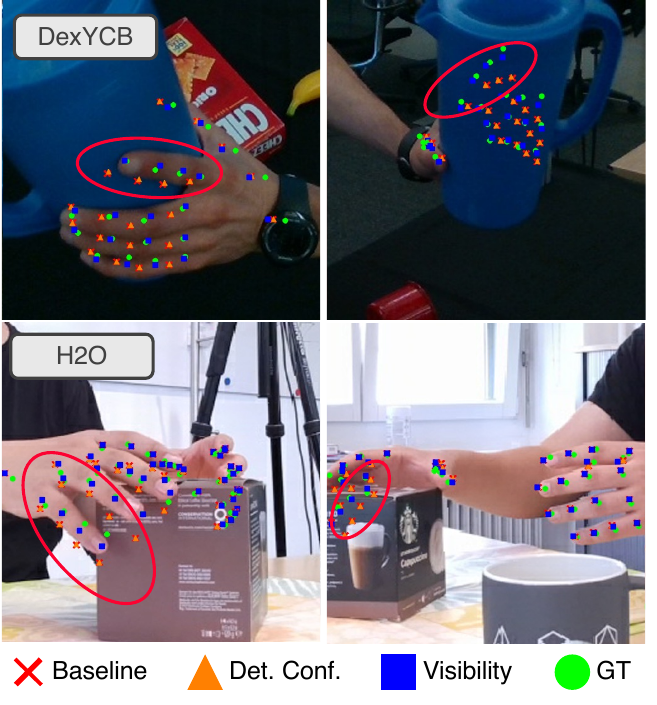}
  \vspace{-2mm}
  \caption{\textbf{Qualitative comparison of triangulation results.} Triangulated 3D joints are reprojected onto each view. Red circles highlight regions where visibility weighting improves the most.}
  \label{fig:triangulation_qual}
  \vspace{-4mm}
\end{figure}

\noindent\textbf{Evaluation.}
\cref{fig:triangulation} shows the results.
Visibility-weighted triangulation achieves lower median, mean, and interquartile range than both the baseline and detection-confidence weighting~\cite{song2026contact4d} on all three datasets.
The improvement is largest on HO3D, which has fewer views and severe occlusion by objects, where the mean reprojection error is reduced by 10.1\%.
These results indicate that per-joint visibility weighting effectively suppresses low-quality 2D estimates of occluded joints.

\cref{fig:triangulation_qual} shows qualitative comparisons for triangulation.
By emphasizing views where each joint is visible, our weighting suppresses 2D estimates displaced by occlusion, reducing reprojection errors.

\vspace{-2mm}
\section{Conclusion}
\label{sec:conclusion}
In this work, we proposed Hand Visibility Detector, the first model dedicated to hand joint visibility estimation.
Leveraging the prior knowledge of large-scale pretrained HPE models, our method substantially outperforms existing methods and general-purpose backbones on the HInt dataset.
Furthermore, visibility-weighted triangulation reduces reprojection errors on multiple datasets.
Future work includes extending our method to video input for temporally consistent visibility estimation.
Our code is released as a ready-to-use package, and we hope it will be utilized in various downstream tasks.\\

\noindent\textbf{Acknowledgment.}
This work was supported by JST K Program JPMJKP25V1.

\clearpage

{
    \small
    \bibliographystyle{ieeenat_fullname}
    \bibliography{main}
}


\end{document}